\documentclass[lettersize,journal]{IEEEtran}
\usepackage{amsmath,amsfonts}
\usepackage{algorithmic}
\usepackage{algorithm}
\usepackage{array}
\usepackage[caption=false,font=normalsize,labelfont=sf,textfont=sf]{subfig}
\usepackage{textcomp}
\usepackage{stfloats}
\usepackage{url}
\usepackage{verbatim}
\usepackage{graphicx}
\usepackage{cite}
\usepackage{amssymb}
\usepackage{booktabs}
\usepackage{float}

\usepackage{xurl}
\usepackage{etoolbox}

\makeatletter
\expandafter\patchcmd\csname\string\IEEEbiography\endcsname
  {\vskip \@IEEEBIOskipN plus 1fil minus 0\baselineskip}
  {\vskip \@IEEEBIOskipN}
  {}
  {\PackageError{BiographySpacing}
    {Cannot patch IEEEbiography}
    {Check the IEEEtran class version.}}
\makeatother

\begin{document}

\title{DualGSL: Prior-Guided Graph Structure Learning with Node-Adaptive Prediction Fusion}

\author{Yuxia Sun,~\IEEEmembership{Member,~IEEE,}
        Yutszyuk Wong,
        Huihong Chen,
        Longhao Lv,
        Chang-Dong Wang,
        and~Li Zeng

\thanks{This work has been submitted to IEEE for possible
publication. Copyright may be transferred without notice,
after which this version may no longer be accessible.}

\thanks{Yuxia Sun is with the College of Information Science
and Technology, Jinan University, Guangzhou 510632, China
(e-mail: tyxsun@email.jnu.edu.cn).}

\thanks{Yutszyuk Wong, Huihong Chen, and Longhao Lv are with
Jinan University, Guangzhou 510632, China
(e-mail: tszyuk1207@gmail.com, 271769167@qq.com,
599844451@qq.com).}

\thanks{Chang-Dong Wang is with Sun Yat-sen University.}

\thanks{Li Zeng is with Changsha University of Science
and Technology.}}

\markboth{}{}

\maketitle

\begin{abstract}
 Graph structure learning (GSL) adapts propagation graphs to prediction tasks, but the relative predictive value of feature-derived priors and learned relations can vary across nodes. We propose DualGSL, a prior-guided framework for transductive node classification that learns node-specific weights for combining predictions from separate prior and task-adaptive propagation paths. The prior branch augments a feature graph with virtual class nodes to incorporate training-label information, while the structural branch combines learned affinity graphs with the feature-graph prior. The two branches share encoder parameters while preserving distinct representations and classification logits. A node-adaptive gate fuses the branch logits based on their projected representations, element-wise absolute difference, and prediction confidence. A joint objective integrates fused and branch-specific classification losses with prior-branch class-anchor alignment, coordinating graph construction, representation learning, and prediction fusion. Experiments against seven baselines on the AMG18 malware
attribution dataset and the widely used 20News text
classification and USPS handwritten-digit recognition
benchmarks show that DualGSL achieves the highest
cross-dataset average accuracy and Macro-F1 among
the evaluated methods. Ablation studies on AMG18 further demonstrate the benefit of combining prior and learned propagation through logit fusion. 
\end{abstract}

\begin{IEEEkeywords}
Graph structure learning, transductive node classification,
prior-guided learning, node-adaptive prediction fusion,
label-enhanced graph learning.
\end{IEEEkeywords}
\raggedbottom
\section{Introduction}
\label{sec:introduction}

\IEEEPARstart{N}{ode} classification is a fundamental task in graph-based learning that supports the organization and analysis of relational data. Graph neural networks (GNNs) address this task by combining node attributes with neighborhood information through message passing~\cite{kipf2017semi,song2023topoenha,yan2025selehop,eliasof2024globloca}, making the propagation structure central to the information used for prediction. When no suitable graph is available, a feature-similarity graph, such as a cosine $k$-nearest-neighbor graph, provides a structural prior derived from node attributes~\cite{chen2020iterative,liu2022compact}. However, feature similarity does not necessarily capture the relations most useful for classification. Graph structure learning (GSL) addresses this mismatch by learning or refining propagation relations to support the prediction task~\cite{chen2020iterative,zhao2023selfsupervised,li2024gagsglob,zheng2024rethstru,han2025unceawar}. A central challenge is therefore to retain useful information from the feature-derived prior while effectively integrating task-adaptive structural information into node prediction.

Existing GSL methods combine prior and learned information at different stages. IDGL incorporates an initial adjacency into learned graph construction~\cite{chen2020iterative}. CoGSL uses node-level prediction confidence to fuse estimated adjacency rows into a propagation graph~\cite{liu2022compact}. ELU-GCN couples representations from the original and learned graphs through a contrastive constraint and combines their GCN outputs before softmax using a scalar coefficient shared across nodes~\cite{huang2024elugcn}. These methods adapt propagation structures or employ a shared output-mixing coefficient. However, the relative predictive value of prior and learned propagation can vary across nodes: learned relations may provide useful additional class evidence for some nodes, while the prior path may be more informative for others. A shared output coefficient applies the same balance to these different cases, motivating a fusion mechanism that learns how the two predictions should contribute to each node.

\begin{figure*}[t] \centering \includegraphics[width=\linewidth]{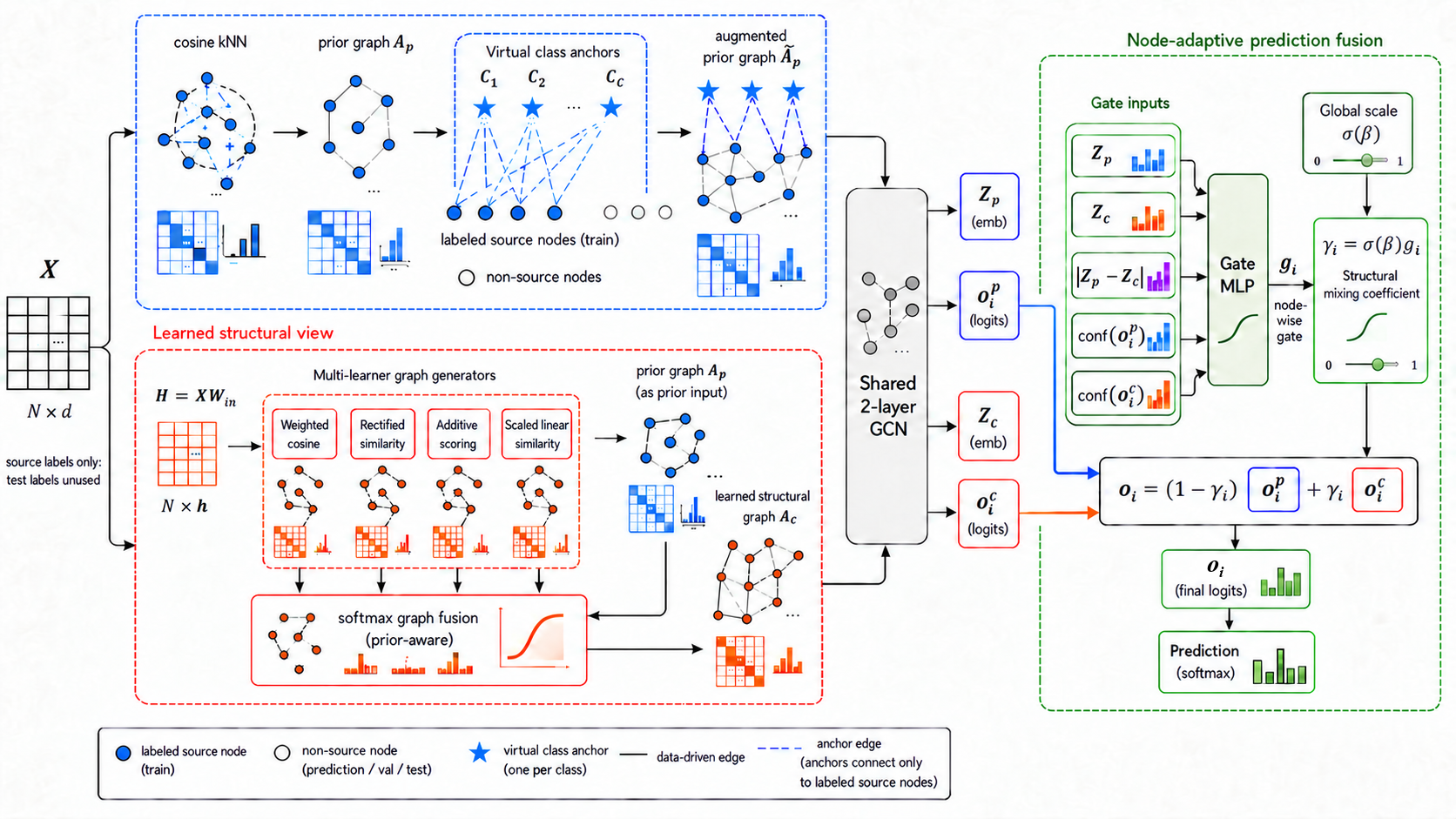} \caption{Overview of the DualGSL architecture.
The label-augmented prior graph
$\widetilde{\mathbf{A}}_p$ and the learned structural graph
$\mathbf{A}_c$ define separate propagation paths with shared
encoder parameters and branch-specific classification heads.
Projected representations, their element-wise absolute difference,
and prediction confidence determine the node-wise gate $g_i$.
The global scale $\sigma(\beta)$ gives the structural mixing
coefficient $\gamma_i=\sigma(\beta)g_i$ for combining the two
branch logits. During training, source-node labels construct
virtual-class connections, while prediction-node labels supervise
the classification and class-anchor alignment objectives.}
\label{fig:framework} \end{figure*}

To this end, we propose \textbf{DualGSL (Dual-view Graph Structure Learning)}, a prior-guided framework for transductive node classification. Our central idea is to retain separate prior and task-adaptive propagation paths and learn node-specific weights for combining their predictions. As illustrated in Fig.~\ref{fig:framework}, the prior branch augments a feature graph with virtual class nodes to incorporate training-label information into message passing. The structural branch constructs four learned affinity graphs from shared hidden features and combines them with the feature-graph prior using global learnable mixture weights. The two branches share encoder parameters while preserving distinct representations and classification logits for node-adaptive prediction fusion.

A node-adaptive gate fuses the branch logits using their projected representations, element-wise absolute difference, and prediction confidence. A learnable global factor scales the node-wise gate, allowing the structural branch to adjust the prior-branch prediction through a node-specific mixing coefficient. Fused and branch-specific classification losses jointly train graph construction, branch representations, and prediction fusion, while class-anchor alignment guides prior-branch node embeddings toward their corresponding propagated virtual class embeddings. The training procedure is summarized in Algorithm~1
(Section~\ref{sec:method}).

We evaluate DualGSL against seven baselines on the AMG18 malware attribution dataset~\cite{amg18dataset} and two widely used benchmarks: 20News for text classification~\cite{lang1995newsweeder} and USPS for handwritten-digit recognition~\cite{hull1994database}. DualGSL achieves the highest mean accuracy and Macro-F1 on AMG18 and USPS and the second-highest mean Macro-F1 on 20News. It also achieves the highest cross-dataset averages among the evaluated methods, with 85.49\% accuracy and 84.62\% Macro-F1. Ablation results on AMG18 further show that DualGSL outperforms the variants without the structural and prior branches by 1.35 and 2.40 percentage points in mean Macro-F1, respectively. Analysis of branch predictions additionally supports the benefit of combining the two propagation paths through logit fusion.

The main contributions are summarized as follows:
\begin{itemize}
    \item We develop a prior-guided GSL framework that
    maintains separate propagation paths over a
    label-augmented feature graph and a task-adaptive
    learned graph, preserving view-specific representations
    and predictions for adaptive prediction-level fusion.

    \item We design a node-adaptive logit fusion mechanism
    conditioned on branch representations, their differences,
    and prediction confidence. A joint learning objective
    integrates fused and branch-specific classification
    with prior-branch class-anchor alignment, coupling
    graph construction, representation learning, and
    prediction fusion.

    \item We demonstrate the effectiveness of DualGSL on
    three datasets, achieving the highest cross-dataset
    average accuracy and Macro-F1 among the evaluated
    methods. Component ablations and fixed-checkpoint
    analyses on AMG18 quantify branch contributions and
    demonstrate the benefit of combining the two
    propagation paths through logit fusion.
\end{itemize}

The remainder of this paper is organized as follows.
Section~2 reviews related work.
Section~3 formulates the transductive node classification problem.
Section~4 presents the DualGSL framework.
Section~5 describes the experimental setup and analyzes the results.
Section~6 concludes the paper and outlines future research directions.

\section{Related Work}

\subsection{Graph Construction and Graph Structure Learning}

GCNs propagate node information over a predefined graph
whose edges remain fixed during training~\cite{kipf2017semi}.
When a suitable graph is unavailable, a feature-similarity
graph provides an initial neighborhood structure, although
feature proximity need not imply usefulness for classification.
Graph structure learning addresses this mismatch by learning
or refining the relations used for message passing~\cite{wu2023nodescal,wang2025weiggrap,han2026infograp}.
IDGL~\cite{chen2020iterative} jointly and iteratively updates
node embeddings and graph structure while incorporating an
initial adjacency. GSR~\cite{zhao2023selfsupervised} instead
uses self-supervised pretraining to refine edges and then
fine-tunes a predictor on the resulting fixed graph.
GaGSL~\cite{li2024gagsglob} combines globally augmented
structural sources under an information-bottleneck objective,
while UnGSL~\cite{han2025unceawar} adjusts directional
connections according to node uncertainty.
However, these structure-learning approaches primarily determine
propagation neighborhoods, whereas DualGSL retains separate
label-augmented prior and task-adaptive propagation paths and
learns how to combine their predictions for each node.

\subsection{Multi-View Graph Learning and Prediction Fusion}

Multi-view graph learning combines information from different
structural sources~\cite{liao2023dualdual,han2023dualadap,wang2024hetegrap,zeng2024multview,shi2025explunif}. CoGSL~\cite{liu2022compact} estimates two
graph views, uses node-level prediction confidence to weight
their adjacency rows, and regularizes the resulting graph
through mutual information compression.
ELU-GCN~\cite{huang2024elugcn} learns a graph to improve label
utilization and applies a contrastive constraint to
representations from the original and learned graphs.
Its final prediction combines the two GCN outputs using
a shared scalar hyperparameter before applying softmax.
In contrast to the confidence-based adjacency fusion of
CoGSL~\cite{liu2022compact} and the shared output-mixing
coefficient of ELU-GCN~\cite{huang2024elugcn}, DualGSL learns
a node-specific logit gate conditioned on branch representations,
their disagreement, and prediction confidence.

\subsection{Label-Aware Graph Learning}

Labels can guide graph learning through class-dependent
relations and explicit label representations~\cite{luo2025labeas,shi2025jlrgcn}.
LRGNN~\cite{liang2023predglob} predicts a global label
relationship matrix through low-rank approximation to support
learning under heterophily. Label-guided graph
rewiring~\cite{bose2025cangrap} uses class embeddings to
reweight similarities and modify neighborhoods.
LEGNN~\cite{yu2023label} represents classes as virtual nodes
and jointly learns node and label representations.
Its training-node selection separates nodes whose labels
construct class connections from nodes whose labels provide
prediction supervision. This prevents a supervised node's
own label from directly determining its class-node connection.
Building on label-node augmentation and training-node
selection~\cite{yu2023label}, DualGSL coordinates a class-aware
prior branch with a task-adaptive structural branch through
node-specific logit fusion, with class-anchor alignment guiding
the prior branch.

\subsection{APT Malware Attribution} \label{sec:rw_apt_attribution}
Advanced persistent threat (APT) malware attribution aims to assign malware samples to their associated APT groups. Learning-based approaches support automated attribution by modeling malware attributes and code patterns. MGAP$^3$ combines statistical features with pretrained code representations through a PerceiverIO architecture~\cite{sun2024mgap}. Related work represents malware using eight types of static features, including byte histograms, byte-entropy histograms, string statistics, and Portable Executable (PE) header information~\cite{Sun2022patentFeature,Sun2023patentAttr,sun2026uap4ma}. AMG18~\cite{amg18dataset}, developed in our previous APT malware attribution research, provides a public dataset for evaluating APT group classification. However, DualGSL takes a complementary perspective by formulating attribution as node classification over feature-derived graphs and jointly learning inter-sample propagation relations and node-adaptive prediction fusion.

\section{Problem Formulation} \label{sec:problem}

We consider transductive node classification from node attributes and partially observed class labels, without assuming a supplied relational graph. Let $\mathcal{V}=\{1,\ldots,N\}$ denote the node set and $\mathbf{X}\in\mathbb{R}^{N\times d}$ the node feature matrix, where $N$ is the number of nodes and $d$ is the feature dimension. The feature vector of node $i$ is $\mathbf{x}_i\in\mathbb{R}^{d}$, with $\mathbf{x}_i^\top$ forming the $i$-th row of $\mathbf{X}$. Each node has a class label $y_i\in\mathcal{Y}=\{1,\ldots,C\}$, where $C$ is the number of classes.

The nodes are divided into pairwise disjoint training, validation, and test sets, denoted by $\mathcal{V}_{\mathrm{tr}}$, $\mathcal{V}_{\mathrm{val}}$, and $\mathcal{V}_{\mathrm{te}}$, respectively, whose union is $\mathcal{V}$. Under the transductive setting, the features of all nodes are available for graph construction and prediction, whereas only training labels are used for parameter optimization and label-aware graph construction. Validation labels are used for model validation, and test labels are reserved for final evaluation.

Given $\mathbf{X}$ and the training labels $\{(i,y_i):i\in\mathcal{V}_{\mathrm{tr}}\}$, the task is to learn a classifier that predicts $\widehat{y}_i\in\mathcal{Y}$ for each test node $i\in\mathcal{V}_{\mathrm{te}}$. Graph-based approaches may construct or learn propagation relations from these available inputs to support classification.

\section{Method}
\label{sec:method}
Our DualGSL retains separate prior and task-adaptive propagation paths and learns  how to combine their predictions for each node. As illustrated in  Fig.~\ref{fig:framework}, the prior branch augments a feature-derived graph with  virtual class nodes to incorporate training-label information into propagation.  The structural branch constructs a learned graph by combining multiple learned  affinity graphs with the feature-graph prior. The two branches share encoder  parameters while producing separate node representations and classification  logits. A node-adaptive gate combines these logits using the projected branch  representations, their element-wise absolute difference, and prediction  confidence to determine the contribution of each branch to the final prediction.  Fused and branch-specific classification losses jointly train graph  construction, representation learning, and prediction fusion. A class-anchor  alignment loss further guides supervised node representations in the prior  branch toward the propagated representations of their corresponding virtual  class nodes.

Throughout the method, the subscripts $p$ and $c$ refer to the prior and structural branches, respectively. We denote the feature-derived prior adjacency matrix by $\mathbf{A}_p$, its label-augmented counterpart containing virtual class nodes by $\widetilde{\mathbf{A}}_p$, and the learned structural adjacency matrix by $\mathbf{A}_c$.
For real nodes, $\mathbf{U}_p$ and $\mathbf{U}_c$ denote the representations produced by the two branches, $\mathbf{Z}_p$ and $\mathbf{Z}_c$ denote their projected representations, and $\mathbf{O}_p$ and $\mathbf{O}_c$ denote their classification logits.

\subsection{Dual-View Graph Construction} \label{sec:graph_construction}

We construct a label-augmented prior graph and a learned structural graph to support the two propagation branches. The branches use a shared hidden feature matrix: \begin{equation} \mathbf{H}=\mathbf{X}\mathbf{W}_{\mathrm{in}}, \end{equation} where $\mathbf{W}_{\mathrm{in}}\in\mathbb{R}^{d\times h}$ is a learnable projection matrix and $h$ is the hidden dimension, giving $\mathbf{H}\in\mathbb{R}^{N\times h}$. The feature-based prior graph is constructed from the input features $\mathbf{X}$, whereas task-adaptive affinities are learned from $\mathbf{H}$.

\subsubsection{Feature-Based Prior Graph}

The prior graph encodes neighborhood relations derived from the input features. Let $\mathcal{N}_k(i)$ denote the $k$ nearest neighbors of node $i$ under cosine distance, excluding node $i$ itself, where $k$ is the neighborhood size. We define a directed affinity matrix $\mathbf{B}\in\mathbb{R}^{N\times N}$ by \begin{equation} B_{ij}= \begin{cases} \max\{0,\cos(\mathbf{x}_i,\mathbf{x}_j)\}, & j\in\mathcal{N}_k(i),\\ 0, & \text{otherwise}, \end{cases} \end{equation} where $\cos(\cdot,\cdot)$ denotes cosine similarity. We symmetrize these affinities to obtain $\mathbf{A}_0$, with entries $A_{0,ij}=\max\{B_{ij},B_{ji}\}$ and a zero diagonal.

For a nonnegative matrix $\mathbf{A}$, define the row-normalization operator $\mathcal{R}$ as \begin{equation} \mathcal{R}(\mathbf{A})_{ij} = \frac{A_{ij}} {\max\{\sum_{\ell}A_{i\ell},\epsilon\}}, \end{equation} where $\epsilon=10^{-12}$ ensures numerical stability. The prior adjacency is $\mathbf{A}_p=\mathcal{R}(\mathbf{A}_0)$ and remains fixed throughout training.

\subsubsection{Label-Augmented Prior Graph}

To incorporate training-label information into
prior-branch propagation, we augment the
feature-derived graph with virtual class nodes. At each training epoch, we randomly partition the training nodes within each class into disjoint source and prediction sets, denoted by $\mathcal{S}$ and $\mathcal{P}$: \begin{equation} \mathcal{S}\cup\mathcal{P} = \mathcal{V}_{\mathrm{tr}}, \qquad \mathcal{S}\cap\mathcal{P} = \varnothing. \end{equation} Each class contributes at least one node to each set. Source-node labels establish edges to virtual class nodes, while prediction-node labels supervise the training objectives. This separation keeps each prediction node's own label from directly determining its virtual-class connection.

We add one virtual node per class, with learnable input embeddings $\mathbf{E}\in\mathbb{R}^{C\times h}$. For class $c$, let $\mathcal{S}_c=\{i\in\mathcal{S}:y_i=c\}$ denote its source nodes. The node-to-class adjacency block $\mathbf{Q}\in\mathbb{R}^{N\times C}$ is defined by \begin{equation} Q_{ic}= \begin{cases} |\mathcal{S}_c|^{-1}, & i\in\mathcal{S}_c,\\ 0, & \text{otherwise}. \end{cases} \end{equation} The augmented feature matrix and adjacency are \begin{equation} \widetilde{\mathbf{H}}_p= \begin{bmatrix} \mathbf{H}\\ \mathbf{E} \end{bmatrix}, \qquad \widetilde{\mathbf{A}}_p= \mathcal{R}\left( \begin{bmatrix} \mathbf{A}_p & \mathbf{Q}\\ \mathbf{Q}^{\top} & \mathbf{0} \end{bmatrix} \right), \label{eq:virtual_graph} \end{equation} where the zero block has size $C\times C$. The bidirectional class-node edges allow virtual nodes to aggregate source-node information and pass class-related information through the prior branch. Although $\mathbf{A}_p$ remains fixed, the augmented adjacency $\widetilde{\mathbf{A}}_p$ changes with the resampled source set and the resulting row normalization. During validation and testing, all training nodes serve as source nodes.

\subsubsection{Learned Structural Graph}

The structural branch learns relations between real nodes using four parameterized affinity functions. Let $\mathbf{h}_i\in\mathbb{R}^{h}$ be the column vector whose transpose forms the $i$-th row of $\mathbf{H}$. The pairwise scores are \begin{align} s_{ij}^{(1)} &= \frac{ (\mathbf{w}\odot\mathbf{h}_i)^{\top} (\mathbf{w}\odot\mathbf{h}_j)} { \|\mathbf{w}\odot\mathbf{h}_i\|_2 \|\mathbf{w}\odot\mathbf{h}_j\|_2}, \\ s_{ij}^{(2)} &= \operatorname{ReLU}(\mathbf{W}_a\mathbf{h}_i)^{\top} \operatorname{ReLU}(\mathbf{W}_a\mathbf{h}_j), \\ s_{ij}^{(3)} &= \operatorname{LeakyReLU}\left( \mathbf{a}^{\top}\mathbf{h}_i+ \mathbf{b}^{\top}\mathbf{h}_j \right), \\ s_{ij}^{(4)} &= \frac{ (\mathbf{W}_m\mathbf{h}_i)^{\top} (\mathbf{W}_m\mathbf{h}_j)} {\sqrt{h}}. \end{align} Here, $\mathbf{w},\mathbf{a},\mathbf{b} \in\mathbb{R}^{h}$ are learnable vectors, $\mathbf{W}_a$ and $\mathbf{W}_m$ are learnable transformation matrices, $\odot$ denotes element-wise multiplication, and $\|\cdot\|_2$ denotes the Euclidean norm. The four functions compute weighted cosine similarity, a dot product after a rectified linear transformation, an additive affinity score, and a scaled dot product after a linear transformation, respectively. Numerical stabilization is applied to cosine normalization.

For each affinity function $r\in\{1,\ldots,4\}$, let $\mathbf{S}^{(r)}\in\mathbb{R}^{N\times N}$ be the score matrix with entries $S_{ij}^{(r)}=s_{ij}^{(r)}$. We obtain a sparse nonnegative adjacency by \begin{equation} \mathbf{A}^{(r)} = \mathcal{T}_k\left( \operatorname{ReLU}(\mathbf{S}^{(r)}) \right), \qquad r=1,\ldots,4, \end{equation} where $\mathcal{T}_k$ retains the $k$ largest entries in each row and sets the remaining entries to zero. The learned adjacencies are not explicitly symmetrized, and diagonal entries are eligible for selection.

The structural adjacency combines the four learned graphs with the feature-derived prior: \begin{equation} \begin{aligned} \mathbf{A}_c &= \alpha_0\mathbf{A}_p+ \sum_{r=1}^{4}\alpha_r \mathcal{R}(\mathbf{A}^{(r)}), \\ \alpha_r &= \frac{\exp(\theta_r)} {\sum_{\ell=0}^{4}\exp(\theta_\ell)}, \qquad r=0,\ldots,4, \end{aligned} \label{eq:graph_fusion} \end{equation} where $\theta_0,\ldots,\theta_4$ are learnable scalar parameters. The resulting global mixture coefficients $\alpha_r$ are shared across nodes and determine the contributions of the prior and learned graphs to structural propagation. Including $\alpha_0\mathbf{A}_p$ preserves prior connectivity while allowing the learned affinities to introduce task-adaptive relations. These coefficients govern graph construction; the node-wise structural mixing coefficients introduced later govern the fusion of branch logits.

\subsection{Shared Graph Encoding} \label{sec:shared_encoding}

The two propagation branches share a two-layer graph encoder. For an input feature matrix $\mathbf{M}$ and an adjacency matrix $\mathbf{A}$, the encoder computes \begin{align} \mathbf{T} &= \operatorname{Dropout}\left( \operatorname{ReLU}\left( \mathbf{A}\mathbf{M}\mathbf{W}_1+\mathbf{b}_1 \right) \right), \\ G_{\theta_{\mathrm{enc}}}(\mathbf{M},\mathbf{A}) &= \mathbf{A}\mathbf{T}\mathbf{W}_2+\mathbf{b}_2. \end{align} Here, $\mathbf{T}$ is the intermediate node representation matrix, and $\theta_{\mathrm{enc}}$ comprises the shared learnable weight matrices $\mathbf{W}_1,\mathbf{W}_2$ and bias vectors $\mathbf{b}_1,\mathbf{b}_2$. Biases are broadcast across nodes.

The encoder output passes through two fully connected layers with LeakyReLU activations, followed by branch-specific batch normalization. The fully connected layers share parameters across branches. Let $F_p$ and $F_c$ denote this transformation with the respective branch-specific normalization. The resulting representations are \begin{align} \mathbf{U}_p^{+} &= F_p\left( G_{\theta_{\mathrm{enc}}}\left( \widetilde{\mathbf{H}}_p, \widetilde{\mathbf{A}}_p \right) \right), \\ \mathbf{U}_c &= F_c\left( G_{\theta_{\mathrm{enc}}}\left( \mathbf{H},\mathbf{A}_c \right) \right). \end{align} Let $d_u$ denote their common representation dimension. Then $\mathbf{U}_p^{+}\in\mathbb{R}^{(N+C)\times d_u}$ contains both real-node and virtual-class-node representations. Its first $N$ rows form $\mathbf{U}_p\in\mathbb{R}^{N\times d_u}$, and its remaining $C$ rows correspond to virtual class nodes. The structural branch produces $\mathbf{U}_c\in\mathbb{R}^{N\times d_u}$ for the real nodes.

Separate linear classification heads produce the branch logits: \begin{align} \mathbf{O}_p &= \mathbf{U}_p\mathbf{W}_p+\mathbf{b}_p, \\ \mathbf{O}_c &= \mathbf{U}_c\mathbf{W}_c+\mathbf{b}_c, \end{align} where $\mathbf{W}_p,\mathbf{W}_c \in\mathbb{R}^{d_u\times C}$ and $\mathbf{b}_p,\mathbf{b}_c\in\mathbb{R}^{C}$ are branch-specific learnable parameters, with biases broadcast across nodes. Both $\mathbf{O}_p$ and $\mathbf{O}_c$ have size $N\times C$.

A shared learnable projection matrix $\mathbf{W}_z\in\mathbb{R}^{d_u\times d_z}$, where $d_z$ is the projection dimension, maps the branch representations into a common space: \begin{equation} \begin{bmatrix} \mathbf{Z}_p\\ \mathbf{Z}_v \end{bmatrix} = \mathbf{U}_p^{+}\mathbf{W}_z, \qquad \mathbf{Z}_c = \mathbf{U}_c\mathbf{W}_z. \end{equation} The projected real-node representations $\mathbf{Z}_p,\mathbf{Z}_c \in\mathbb{R}^{N\times d_z}$ provide inputs to the node-adaptive fusion mechanism. The matrix $\mathbf{Z}_v\in\mathbb{R}^{C\times d_z}$ contains the virtual class representations after propagation and projection, which serve as class anchors for the alignment objective.

\subsection{Node-Adaptive Logit Fusion} \label{sec:adaptive_fusion}

DualGSL combines the branch predictions using a mixing coefficient adapted to each node. For real node $i$, let $\mathbf{o}_i^p,\mathbf{o}_i^c\in\mathbb{R}^{C}$ and $\mathbf{z}_i^p,\mathbf{z}_i^c\in\mathbb{R}^{d_z}$ denote its branch logits and projected representations, respectively. These are column vectors whose transposes form the corresponding rows of $\mathbf{O}_p,\mathbf{O}_c$ and $\mathbf{Z}_p,\mathbf{Z}_c$. For each branch $v\in\{p,c\}$, we compute a prediction confidence score as the maximum softmax probability: \begin{equation} q_i^v = \max_{1\leq \ell\leq C} \left[ \operatorname{softmax}(\mathbf{o}_i^v) \right]_{\ell}. \end{equation}

The gate input combines the projected branch representations, their element-wise absolute difference, and the two confidence scores: \begin{equation} \mathbf{r}_i = [ \mathbf{z}_i^p \Vert\mathbf{z}_i^c \Vert|\mathbf{z}_i^p-\mathbf{z}_i^c| \Vert q_i^p \Vert q_i^c ], \end{equation} where $\Vert$ denotes concatenation and $|\cdot|$ denotes element-wise absolute value, giving $\mathbf{r}_i\in\mathbb{R}^{3d_z+2}$. The representations describe the information available from each branch, their difference characterizes branch disagreement, and the confidence scores summarize the respective predictions.

A learnable gating network produces the node-wise gate $g_i$, which is scaled to obtain the structural mixing coefficient $\gamma_i$: \begin{equation} g_i = \sigma\left( \operatorname{MLP}_g(\mathbf{r}_i) \right), \qquad \gamma_i = \sigma(\beta)g_i. \end{equation} Here, $\sigma$ is the sigmoid function, $\operatorname{MLP}_g$ consists of two linear layers with an intermediate LeakyReLU activation and a scalar output, and $\beta$ is a learnable global scalar. The global scale $\sigma(\beta)$ sets a shared upper bound on the structural mixing coefficients, while $g_i$ adapts the coefficient to each node. Both the gating network and the global scale are learned jointly with graph construction and the branch predictors.

The fused logit vector is \begin{equation} \mathbf{o}_i = (1-\gamma_i)\mathbf{o}_i^p + \gamma_i\mathbf{o}_i^c. \label{eq:node_fusion} \end{equation} Equivalently, \begin{equation} \mathbf{o}_i = \mathbf{o}_i^p + \gamma_i \left( \mathbf{o}_i^c-\mathbf{o}_i^p \right). \end{equation} This form makes the prior-guided prediction mechanism explicit: the prior-branch logits provide a reference, and the difference between the branch logits supplies a node-specific adjustment controlled by $\gamma_i$. We initialize $\beta=-2$ to favor the prior branch at the start of training, while allowing the global scale and node-wise gate to adapt during optimization. The final predicted label is \begin{equation} \widehat{y}_i = \arg\max_{1\leq \ell\leq C} o_{i\ell}, \end{equation} where $o_{i\ell}$ is the $\ell$-th element of the fused logit vector $\mathbf{o}_i$.

\subsection{Joint Objective and Training} \label{sec:objectives}

\subsubsection{Classification Supervision}

We supervise the fused predictions on the prediction set $\mathcal{P}$ using cross-entropy: \begin{equation} \mathcal{L}_{\mathrm{cls}} = -\frac{1}{|\mathcal{P}|} \sum_{i\in\mathcal{P}} \log \left[ \operatorname{softmax}(\mathbf{o}_i) \right]_{y_i}. \end{equation} We also apply classification supervision directly to both branches: \begin{equation} \mathcal{L}_{\mathrm{view}} = -\frac{1}{|\mathcal{P}|} \sum_{i\in\mathcal{P}} \sum_{v\in\{p,c\}} \log \left[ \operatorname{softmax}(\mathbf{o}_i^v) \right]_{y_i}. \end{equation} The fused loss trains the combined predictor, while the branch-specific loss provides a direct classification signal to each propagation path.

\subsubsection{Class-Anchor Alignment}

The propagated virtual class representations provide class anchors for the prior branch. Let $\mathbf{z}_{v,\ell}\in\mathbb{R}^{d_z}$ denote the column vector whose transpose forms the $\ell$-th row of $\mathbf{Z}_v$. For each supervised prediction node $i\in\mathcal{P}$, we align its projected prior-branch representation with the anchor of its class: \begin{equation} \mathcal{L}_{\mathrm{center}} = \frac{1}{|\mathcal{P}|d_z} \sum_{i\in\mathcal{P}} \left\| \overline{\mathbf{z}}_i^p - \overline{\mathbf{z}}_{v,y_i} \right\|_2^2, \end{equation} where the overline denotes $\ell_2$ normalization. The denominator averages the squared error over prediction nodes and projection coordinates. This objective encourages prediction nodes of the same class to align with their shared virtual class anchor in the projected prior-branch space.

\begin{algorithm}[H]
\caption{DualGSL Training and Inference}\label{alg:dualgsl}
\begin{algorithmic}[1]

\REQUIRE Features $\mathbf{X}$ of all nodes,
training and validation labels, and node splits;
model configuration
$m\in\{\mathrm{dual},\mathrm{prior}\}$;
neighborhood size $k$, source ratio, loss weights,
Adam hyperparameters, maximum epochs $T$,
patience $p$, and minimum improvement $\delta$.
\ENSURE Selected model state $\Theta^{*}$
and test predictions under configuration $m$.

\STATE Construct the nonnegative, symmetrized
cosine $k$-nearest-neighbor adjacency
$\mathbf{A}_0$ from $\mathbf{X}$.
\STATE Set
$\mathbf{A}_p=\mathcal{R}(\mathbf{A}_0)$.
\STATE Initialize the model under configuration $m$.
\IF{$m=\mathrm{dual}$}
    \STATE Initialize the global scaling parameter
    $\beta=-2$.
\ENDIF
\STATE Set validation reference score $b=-\infty$
and consecutive non-improvement counter $q=0$.

\FOR{$t=1,\ldots,T$}
    \STATE Enable training mode.
    \STATE Randomly partition
    $\mathcal{V}_{\mathrm{tr}}$ within each class
    into source nodes $\mathcal{S}$ and prediction
    nodes $\mathcal{P}$ using the specified source ratio.
    \STATE Compute
    $\mathbf{H}=\mathbf{X}\mathbf{W}_{\mathrm{in}}$.
    \STATE Construct the label-augmented prior
    features and adjacency using $\mathcal{S}$
    via Eq.~\eqref{eq:virtual_graph}.
    \STATE Encode
    $(\widetilde{\mathbf{H}}_p,
      \widetilde{\mathbf{A}}_p)$
    and compute $\mathbf{O}_p$, $\mathbf{Z}_p$,
    and $\mathbf{Z}_v$.

    \IF{$m=\mathrm{dual}$}
        \STATE Construct the four learned affinity
        graphs and combine them with $\mathbf{A}_p$
        to obtain $\mathbf{A}_c$
        via Eq.~\eqref{eq:graph_fusion}.
        \STATE Encode $(\mathbf{H},\mathbf{A}_c)$
        and compute $\mathbf{O}_c$ and $\mathbf{Z}_c$.
        \STATE Compute the node-wise gates $g_i$
        and structural mixing coefficients $\gamma_i$;
        fuse the branch logits using
        Eq.~\eqref{eq:node_fusion}.
        \STATE Compute the dual-view loss
        on $\mathcal{P}$ using
        Eq.~\eqref{eq:total_loss}.
    \ELSE
        \STATE Compute the prior-only loss
        on $\mathcal{P}$ using
        Eq.~\eqref{eq:prior_loss}.
    \ENDIF

    \STATE Update trainable parameters
    with one Adam step.
    \STATE Enable evaluation mode
    without gradient computation.
    \STATE Recompute the full forward pass,
    including graph construction and prediction,
    under configuration $m$ using the updated
    model state and
    $\mathcal{S}=\mathcal{V}_{\mathrm{tr}}$.
    \STATE Compute validation Macro-F1 $f_t$
    on $\mathcal{V}_{\mathrm{val}}$.

    \IF{$f_t>b+\delta$}
        \STATE Save the current model state
        as $\Theta_{\mathrm{best}}$.
        \STATE Set $b\leftarrow f_t$
        and $q\leftarrow0$.
    \ELSE
        \STATE Set $q\leftarrow q+1$.
    \ENDIF

    \IF{$q\geq p$}
        \STATE \textbf{break}
    \ENDIF
\ENDFOR

\STATE Set
$\Theta^{*}\leftarrow\Theta_{\mathrm{best}}$.
\STATE Restore $\Theta^{*}$ and enable evaluation
mode without gradient computation.
\STATE Recompute the full forward pass under
configuration $m$ using
$\mathcal{S}=\mathcal{V}_{\mathrm{tr}}$,
and output predictions for
$\mathcal{V}_{\mathrm{te}}$.

\end{algorithmic}
\end{algorithm}

\subsubsection{Joint Learning Objective}

The dual-view objective combines fused classification, branch-specific classification, and class-anchor alignment: \begin{equation}
\mathcal{L} = \mathcal{L}_{\mathrm{cls}} + \lambda_{\mathrm{view}} \mathcal{L}_{\mathrm{view}} + \lambda_{\mathrm{center}} \mathcal{L}_{\mathrm{center}}, \label{eq:total_loss} 
\end{equation} where $\lambda_{\mathrm{view}}\geq 0$ and $\lambda_{\mathrm{center}}\geq 0$ control the contributions of branch-specific supervision and class-anchor alignment, respectively. Together, these objectives jointly optimize graph construction, branch representations, and node-adaptive prediction fusion.

For the prior-only configuration, we independently initialize and train the label-augmented prior branch using \begin{equation} \mathcal{L}_{\mathrm{prior}} = \mathcal{L}_{\mathrm{cls}}^{p} + \lambda_{\mathrm{center}} \mathcal{L}_{\mathrm{center}}, \label{eq:prior_loss} \end{equation} where \begin{equation} \mathcal{L}_{\mathrm{cls}}^{p} = -\frac{1}{|\mathcal{P}|} \sum_{i\in\mathcal{P}} \log \left[ \operatorname{softmax}(\mathbf{o}_i^p) \right]_{y_i}. \end{equation} This configuration retains virtual class nodes and class-anchor alignment and predicts labels directly from the prior-branch logits.

\subsubsection{Training Procedure}

Let $m\in\{\mathrm{dual},\mathrm{prior}\}$ denote the model configuration specified for a training run. Algorithm~1 optimizes the corresponding objective with Adam, using Eq.~\eqref{eq:total_loss} for $m=\mathrm{dual}$ and Eq.~\eqref{eq:prior_loss} for $m=\mathrm{prior}$. At each epoch, the training nodes are randomly partitioned within each class into source and prediction sets. Source-node labels construct the virtual class connections, and all classification and alignment losses are evaluated on the prediction nodes.

After each parameter update, validation predictions are recomputed in evaluation mode, with all training nodes serving as source nodes. Let $f_t$ denote validation Macro-F1 at epoch $t$, and let $b$ denote the score associated with the saved checkpoint. The checkpoint is updated when $f_t>b+\delta$, where $\delta\geq 0$ is the prescribed minimum improvement. Training stops after $p$ consecutive epochs without such an update or after the maximum of $T$ epochs, where $p$ is the patience. The saved model state is then restored for final test prediction, again using all training nodes as sources. The same configuration $m$ is used throughout training, validation, and testing.

\section{Experiments}
\label{sec:experiments}

We evaluate DualGSL and seven competitor methods on three datasets. 

\subsection{Experimental Setup}
\label{sec:experimental_setup}

\subsubsection{Datasets and Evaluation Protocol}
The three datasets used in our experiments are AMG18 for malware attribution,
20News for text classification~\cite{lang1995newsweeder},
and USPS for handwritten-digit
recognition~\cite{hull1994database}.

\textbf{AMG18.} AMG18~\cite{amg18dataset} contains 6,360 malware samples from 18 APT groups. It extends the APTMalware collection~\cite{aptmalware2019} with additional hash--label pairs collected from public threat repositories, with the corresponding malware samples obtained from VirusSign~\cite{virussign}. Each sample corresponds to a node, and its APT group label defines the classification target.

\textbf{AMG18 feature representation.} Using the eight static feature types summarized in Section~\ref{sec:rw_apt_attribution}, we encode each malware sample as a 2,351-dimensional vector. Principal component analysis (PCA) reduces these vectors to 128 dimensions. The resulting vectors serve as node attributes for model input and for constructing the cosine $k$-nearest-neighbor prior graph.

For 20News and USPS, we extract features using the
\texttt{all-mpnet-base-v2} sentence encoder~\cite{allmpnetbasev2}
and an ImageNet-pretrained ResNet-18~\cite{he2016deep},
respectively. The resulting representations, with their
final dimensions reported in Table~1, serve as node
attributes for classification and feature-based graph
construction.

\begin{table}[t]
\centering
\caption{Dataset statistics after validation splitting.}
\label{tab:dataset_statistics}
\small
\setlength{\tabcolsep}{4pt}
\begin{tabular}{lcccccc}
\toprule
Dataset & Nodes & Features & Classes & Train & Val & Test \\
\midrule
AMG18 & 6,360 & 128 & 18 & 4,573 & 509 & 1,278 \\
20News & 18,846 & 768 & 20 & 10,182 & 1,132 & 7,532 \\
USPS & 9,298 & 256 & 10 & 6,561 & 730 & 2,007 \\
\bottomrule
\end{tabular}
\end{table}

Table~\ref{tab:dataset_statistics} summarizes each dataset's node count, input feature dimensionality after preprocessing, number of classes, and training, validation, and test split sizes. For each dataset, we retain the provided test split and reserve 10\% of the original training partition for validation using stratified sampling. The remaining training nodes are used for model optimization. Under the transductive setting, node features from all three partitions are available for graph construction, while only training labels are used for label-aware graph construction and parameter optimization.

At each training epoch, training nodes are randomly partitioned within each class into disjoint source and prediction sets, with a source ratio of 0.5. Source-node labels establish connections to virtual class nodes, while prediction-node labels supervise the training objectives. During validation and testing, all training nodes serve as source nodes. Following Algorithm~1, we use validation Macro-F1
for checkpoint selection and early stopping within
each run. The selected checkpoint is used for
final test evaluation.

\subsubsection{Baselines and Metrics}

We compare DualGSL with seven baselines covering feature-only classification, fixed-graph propagation, graph structure learning, and label-enhanced learning. MLP predicts labels directly from node features, providing a reference for assessing the benefit of graph-based modeling. GCN~\cite{kipf2017semi} aggregates neighborhood information through graph convolutions over a preconstructed graph that remains fixed during training.

GSR~\cite{zhao2023selfsupervised} uses self-supervised contrastive pretraining to refine graph edges, followed by supervised fine-tuning on the fixed refined graph. IDGL~\cite{chen2020iterative} jointly and iteratively learns graph structure and node representations. ELU-GCN~\cite{huang2024elugcn} learns a graph to enhance label utilization and applies contrastive learning to representations from the original and learned graphs. CoGSL~\cite{liu2022compact} adaptively fuses estimated graph views into a propagation graph and uses mutual information compression to reduce redundancy. LEGNN~\cite{yu2023label} introduces virtual class nodes and jointly learns node and label representations to incorporate label information into propagation.

We report accuracy and Macro-F1 as percentages. Accuracy measures the proportion of correctly classified test nodes, while Macro-F1 is the unweighted arithmetic mean of the class-wise F1 scores. By assigning equal weight to each class, Macro-F1 complements overall accuracy when class distributions are imbalanced.
Unless otherwise specified, experiments are repeated with five random seeds, and mean scores are reported. Where shown, $\pm$ denotes the standard deviation across these runs. For each metric, the cross-dataset average in Table~\ref{tab:main_results} is the unweighted arithmetic mean of the three per-dataset mean scores.

All compared methods use the same node features and
training, validation, and test partitions for each dataset.
Graph-based baselines that require an initial graph use a
feature-derived graph constructed from these inputs.
Each baseline retains its original learning objective,
and its hyperparameters and checkpoint are determined
using the validation set. Test labels are used only for
final evaluation.

\begin{table*}[t!]
\centering
\caption{Dataset-specific hyperparameters used for
the main results. Each model is trained using
Algorithm~1 with five random seeds.
A dash indicates an inactive term.}
\label{tab:hyperparameters}
\small
\setlength{\tabcolsep}{5pt}
\begin{tabular}{lccccc}
\toprule
Dataset & $k$ & Source ratio
& $\lambda_{\mathrm{center}}$
& $\lambda_{\mathrm{view}}$
& Seeds \\
\midrule
AMG18 & 10 & 0.5 & 0.05 & 0.40 & 1--5 \\
20News & 5 & 0.5 & 0.05 & 0.05 & 1--5 \\
USPS & 3 & 0.5 & 0.05 & -- & 1--5 \\
\bottomrule
\end{tabular}
\end{table*}

\subsubsection{Implementation Details}

We implement DualGSL in PyTorch and optimize it using Adam with a learning rate of 0.005 and weight decay of $5\times10^{-4}$. The shared graph encoder has two layers and uses a dropout rate of 0.2. For the dual-view model, the global fusion parameter $\beta$ is initialized to $-2$.

Each run follows Algorithm~1 for at most 1,000 epochs. Validation Macro-F1 determines checkpoint selection and early stopping, with dataset-specific patience and minimum improvement thresholds. Table~\ref{tab:hyperparameters} summarizes the dataset-specific hyperparameters and random seeds used for the main results.

On AMG18 and 20News, we use the dual-view configuration. On USPS, the prior-only configuration is trained independently using the objective in Eq.~\eqref{eq:prior_loss}. It retains label-augmented prior propagation and class-anchor alignment and uses the prior-branch logits for prediction. For each dataset, the specified configuration is used for both training and test evaluation.

\begin{table*}[t] 
\centering 
\caption{Classification performance (\%). Per-dataset results are reported as mean $\pm$ standard deviation over five random seeds. The last two columns report the unweighted averages of the three per-dataset means. Bold and underlined values indicate the best and second-best means in each column, respectively. }
 
\label{tab:main_results} \small \setlength{\tabcolsep}{3.5pt} \renewcommand{\arraystretch}{1.08}

\begin{tabular}{@{}lcccccccc@{}} \toprule & \multicolumn{2}{c}{AMG18} & \multicolumn{2}{c}{20News} & \multicolumn{2}{c}{USPS} & \multicolumn{2}{c}{Cross-dataset average} \\ \cmidrule(lr){2-3} \cmidrule(lr){4-5} \cmidrule(lr){6-7} \cmidrule(lr){8-9} Method & Acc. & Macro-F1 & Acc. & Macro-F1 & Acc. & Macro-F1 & Avg. Acc. & Avg. Macro-F1 \\ \midrule

MLP & $\underline{88.65}\pm0.31$ & $\underline{86.85}\pm0.72$ & $71.94\pm0.13$ & $70.66\pm0.08$ & $93.16\pm0.10$ & $92.61\pm0.14$ & $84.58$ & $83.37$ \\

GCN~\cite{kipf2017semi} & $88.29\pm0.14$ & $86.29\pm0.11$ & $\mathbf{73.59}\pm0.13$ & $\mathbf{72.30}\pm0.12$ & $93.54\pm0.21$ & $93.13\pm0.22$ & $85.14$ & $83.91$ \\

GSR~\cite{zhao2023selfsupervised} & $\underline{88.65}\pm0.21$ & $86.68\pm0.31$ & $72.48\pm0.27$ & $71.32\pm0.29$ & $94.24\pm0.19$ & $93.77\pm0.21$ & $85.12$ & $\underline{83.92}$ \\

IDGL~\cite{chen2020iterative} & $82.46\pm0.59$ & $77.98\pm1.07$ & $70.03\pm0.27$ & $68.56\pm0.38$ & $91.66\pm0.30$ & $90.92\pm0.36$ & $81.38$ & $79.15$ \\

ELU-GCN~\cite{huang2024elugcn} & $86.90\pm0.27$ & $84.51\pm0.26$ & $\underline{72.92}\pm0.05$ & $71.44\pm0.18$ & $93.38\pm0.24$ & $92.99\pm0.26$ & $84.40$ & $82.98$ \\

CoGSL~\cite{liu2022compact} & $86.60\pm0.07$ & $84.35\pm0.18$ & $65.75\pm0.30$ & $64.36\pm0.19$ & $92.89\pm0.12$ & $92.44\pm0.12$ & $81.75$ & $80.38$ \\

LEGNN~\cite{yu2023label} & $88.23\pm0.38$ & $85.26\pm0.90$ & $72.15\pm0.32$ & $71.06\pm0.34$ & $\underline{95.17}\pm0.23$ & $\underline{94.79}\pm0.25$ & $\underline{85.18}$ & $83.70$ \\

\midrule

DualGSL & $\mathbf{88.76}\pm0.23$ & $\mathbf{87.28}\pm0.64$ & $72.35\pm1.82$ & $\underline{71.58}\pm1.43$ & $\mathbf{95.36}\pm0.15$ & $\mathbf{95.00}\pm0.13$ & $\mathbf{85.49}$ & $\mathbf{84.62}$ \\

\bottomrule \end{tabular} 
\end{table*}

\subsection{Classification Performance} \label{sec:comparison_results}

Table~\ref{tab:main_results} compares DualGSL with seven baselines across the three datasets. DualGSL achieves the highest cross-dataset average accuracy (85.49\%) and Macro-F1 (84.62\%). These exceed the strongest baseline averages, obtained by LEGNN for accuracy and GSR for Macro-F1, by 0.31 and 0.70 percentage points, respectively.

\subsubsection{Malware Attribution on AMG18}

On AMG18, DualGSL achieves the highest mean accuracy (88.76\%) and Macro-F1 (87.28\%) among the evaluated methods. These results support the effectiveness of the proposed framework for APT group attribution from static malware features. The subsequent ablation studies and structural analyses examine the contributions of the two propagation paths and their prediction fusion.

\subsubsection{Text Classification on 20News}

On 20News, DualGSL achieves a mean accuracy of 72.35\% and a mean Macro-F1 of 71.58\%. It ranks second in Macro-F1, following GCN (72.30\%), and achieves a higher score on this metric than the evaluated graph structure learning baselines.

\subsubsection{Handwritten-Digit Recognition on USPS}

On USPS, the independently trained prior-only configuration achieves the highest mean accuracy (95.36\%) and Macro-F1 (95.00\%) among the evaluated methods. These scores exceed those of LEGNN, the strongest baseline on this dataset, by 0.19 and 0.21 percentage points, respectively.

\subsection{Ablation Study} \label{sec:ablation}

We conduct ablation experiments on the AMG18 dataset to examine the contributions of the dual-branch architecture and its components. Table~\ref{tab:ablation_apt} compares the full model with variants that remove individual propagation branches, graph learners, virtual class nodes, or class-anchor alignment. The two branch-removal variants are trained independently using the remaining propagation branch.

\begin{table}[t] \centering \caption{Ablation study on AMG18. Results are averaged over five random seeds.} \label{tab:ablation_apt} \small \setlength{\tabcolsep}{3pt} \begin{tabular}{@{}lrrrr@{}} \toprule Variant & \shortstack{Acc.\\(\%)} & \shortstack{$\Delta$Acc.\\(pp)} & \shortstack{Macro-F1\\(\%)} & \shortstack{$\Delta$Macro-F1\\(pp)} \\ \midrule Full model & 88.76 & --- & 87.28 & --- \\ w/o structural branch & 87.87 & $-0.89$ & 85.93 & $-1.35$ \\ w/o prior branch & 87.37 & $-1.39$ & 84.88 & $-2.40$ \\ \midrule w/o graph learners & 88.09 & $-0.67$ & 86.04 & $-1.24$ \\ w/o virtual class nodes & 88.56 & $-0.20$ & 86.88 & $-0.40$ \\ w/o center alignment & 88.59 & $-0.17$ & 86.98 & $-0.30$ \\ \bottomrule \end{tabular}

\smallskip \begin{minipage}{\columnwidth} \footnotesize \textit{Note:} $\Delta$ denotes each variant's mean minus the full model's mean, measured in percentage points (pp). Negative values indicate a decrease. \end{minipage} \end{table}

The full model achieves the highest mean accuracy and Macro-F1 among the evaluated variants. Removing the structural branch reduces Macro-F1 by 1.35 percentage points and accuracy by 0.89 points, demonstrating the benefit of incorporating learned structural propagation alongside the label-augmented prior branch. Removing the prior branch reduces Macro-F1 by 2.40 points and accuracy by 1.39 points, highlighting the value of retaining an explicit prior propagation path. Together, these results support the combined use of the two branches for prediction.

Within the dual-branch architecture, removing the graph learners decreases Macro-F1 by 1.24 points and accuracy by 0.67 points, supporting the contribution of task-adaptive graph construction. Removing virtual class nodes or class-anchor alignment further lowers Macro-F1 by 0.40 and 0.30 points, respectively, with corresponding decreases in accuracy. These results show that label-aware propagation and class-anchor alignment further improve the dual-branch predictor.

\subsection{Prediction and Structural Analysis} \label{sec:apt_structure_analysis}
We examine the predictive benefit of branch fusion, the edges introduced by structure learning, and the representations and mixing coefficients produced by DualGSL on AMG18. The quantitative prediction and edge analyses use the checkpoint selected by validation Macro-F1 from the seed-1 run under the AMG18 configuration in Table~\ref{tab:hyperparameters}.

\subsubsection{Prediction Fusion Analysis} \label{sec:amg18_prediction_fusion}

We compare predictions obtained from the prior-branch logits, structural-branch logits, and fused logits of the same trained dual-view model. Table~\ref{tab:apt_prediction_diagnostics} reports their test performance at the seed-1 checkpoint specified above.

\begin{table}[t] 
\centering 
\caption{Prediction fusion on AMG18 at the same trained dual-view checkpoint. Bold values indicate the best results.} \label{tab:apt_prediction_diagnostics} \small \setlength{\tabcolsep}{5pt} \begin{tabular}{lcc} \toprule Prediction & Acc. (\%) & Macro-F1 (\%) \\ \midrule Prior branch & 87.72 & 86.52 \\ Structural branch & 87.32 & 85.95 \\ Fused prediction & \textbf{88.73} & \textbf{88.16} \\ \bottomrule \end{tabular} 
\end{table}

As Table~\ref{tab:apt_prediction_diagnostics} shows, the fused prediction achieves higher accuracy and Macro-F1 than either branch prediction, improving Macro-F1 over the prior and structural heads by 1.64 and 2.21 percentage points, respectively. Relative to the prior-head prediction, fusion corrects 15 test-node errors and introduces two errors, yielding 13 additional correct predictions. These results demonstrate that combining the branch logits improves classification beyond using either head alone.

At the seed-1 checkpoint evaluated in Table~\ref{tab:apt_prediction_diagnostics}, the mean structural mixing coefficient $\gamma_i$ over all test nodes is 19.28\%. The predictive gains are achieved even though the logit mixture assigns greater weight to the prior branch on average. Together with the branch prediction comparison, this result supports prior-guided prediction fusion: the learned structural branch provides useful logit adjustments that improve the prior-branch prediction.

\subsubsection{Learned Graph Structure Analysis} \label{sec:amg18_learned_structure}

We examine how the learned structural graph $\mathbf{A}_c$ expands the feature-derived prior graph $\mathbf{A}_p$ and how the new edges introduced in $\mathbf{A}_c$ affect prediction. The statistics cover positive-weight directed edges between real malware-sample nodes, excluding self-loops and edges involving virtual class nodes.

We partition the edges included in the statistics for $\mathbf{A}_c$ into two subsets: \textit{\textbf{ Inherited edges}} are those also present in $\mathbf{A}_p$, whereas\textit{\textbf{ new edges}} are absent from $\mathbf{A}_p$. Thus, the inherited edges are the common edges of the two graphs, and the two subsets together comprise all edges counted for $\mathbf{A}_c$. Table~\ref{tab:apt_graph_structure} reports the statistics for both graphs and the two edge subsets of $\mathbf{A}_c$. Class labels are used to characterize these edges post hoc.

\begin{table}[t] \centering \caption{Edge counts and class consistency on AMG18. \# Edges denotes the number of positive-weight, non-self directed edges between real nodes. SC denotes same-class. For each row, SC edges reports the percentage of edges joining nodes with the same class label, and SC weight reports the percentage of total edge weight carried by these same-class edges. Weights are taken from $\mathbf{A}_p$ for the first row and from $\mathbf{A}_c$ for the remaining rows.} \label{tab:apt_graph_structure} \small \setlength{\tabcolsep}{4pt} \begin{tabular}{@{}lrrr@{}} \toprule Graph / subset & \# Edges & \shortstack{SC edges\\(\%)} & \shortstack{SC weight\\(\%)} \\ \midrule Prior graph $\mathbf{A}_p$ & 89,756 & 93.21 & 93.09 \\ Structural graph $\mathbf{A}_c$ & 283,647 & 53.44 & 68.68 \\ Inherited edges in $\mathbf{A}_c$ & 89,756 & 93.21 & 95.68 \\ New edges in $\mathbf{A}_c$ & 193,891 & 35.04 & 41.24 \\ \bottomrule \end{tabular} \end{table}

The inherited-edge count in $\mathbf{A}_c$
equals the prior-graph edge count (89,756),
confirming that all prior edges are retained.
As shown in Table~\ref{tab:apt_graph_structure},
$\mathbf{A}_c$ additionally contains 193,891
new edges.
These constitute 68.36\% of its 283,647
non-self edges
($193{,}891/283{,}647 \approx 68.36\%$).
The structural graph therefore preserves
the prior connectivity while adding neighbors
beyond those supplied by the prior graph.

Table~\ref{tab:apt_graph_structure} also shows that same-class edges account for 93.21\% of the edges in $\mathbf{A}_p$ and 35.04\% of the new edges in $\mathbf{A}_c$. Within the new-edge subset, same-class edges carry 41.24\% of the total weight. These count and weight shares imply that same-class new edges have an average weight approximately 1.30 times that of cross-class new edges. Thus, the structural graph expands the prior neighborhoods with a larger proportion of cross-class edges, while assigning greater average propagation weight to same-class new edges.

Furthermore, using the checkpoint in Table~\ref{tab:apt_prediction_diagnostics}, we compare removing all new edges from $\mathbf{A}_c$ with random removal that matches the number of deleted non-self edges in each row (random seed 1001). In both conditions, self-loops are retained and adjacency rows are renormalized. Model parameters and the prior branch remain fixed, while structural representations and mixing coefficients are recomputed.

\begin{table}[t] \centering \caption{Edge-removal analysis on AMG18 at a fixed checkpoint. Removed weight is the percentage of the full $\mathbf{A}_c$'s non-self edge weight removed, computed before deletion and renormalization. Bold values indicate the best prediction results.} \label{tab:amg18_connection_removal} \small \setlength{\tabcolsep}{3pt} \begin{tabular}{@{}lccc@{}} \toprule Graph condition & \shortstack{Removed\\weight (\%)} & \shortstack{Acc.\\(\%)} & \shortstack{Macro-F1\\(\%)} \\ \midrule Full $\mathbf{A}_c$ & 0.00 & 88.73 & \textbf{88.16} \\ New edges removed & 49.59 & \textbf{88.97} & 88.01 \\ Random edges removed & 69.14 & 88.58 & 87.65 \\ \bottomrule \end{tabular} \end{table}

Table~\ref{tab:amg18_connection_removal} shows that removing new edges decreases Macro-F1 by 0.15 percentage points and increases accuracy by 0.24 percentage points. Random removal lowers both metrics relative to the full graph and removes more edge weight than new-edge removal (69.14\% versus 49.59\%), despite matching the per-row deletion counts.
After new-edge removal, the fused prediction still achieves 88.01\% Macro-F1, exceeding the prior-head result of 86.52\% in Table~\ref{tab:apt_prediction_diagnostics}. Thus, at this checkpoint, fusion retains a predictive benefit over the prior head even when the structural graph contains only inherited edges and self-loops. Retaining the new edges further improves Macro-F1, giving the full graph the highest score on this metric among the evaluated conditions.

In summary, the learned structural graph preserves prior connectivity while expanding propagation neighborhoods. At the examined checkpoint, fusion retains a Macro-F1 advantage over the prior head after new-edge removal, while retaining the full structural graph achieves the highest Macro-F1 among the evaluated conditions.

\begin{figure*}[t]
\centering \includegraphics[width=0.95\linewidth] {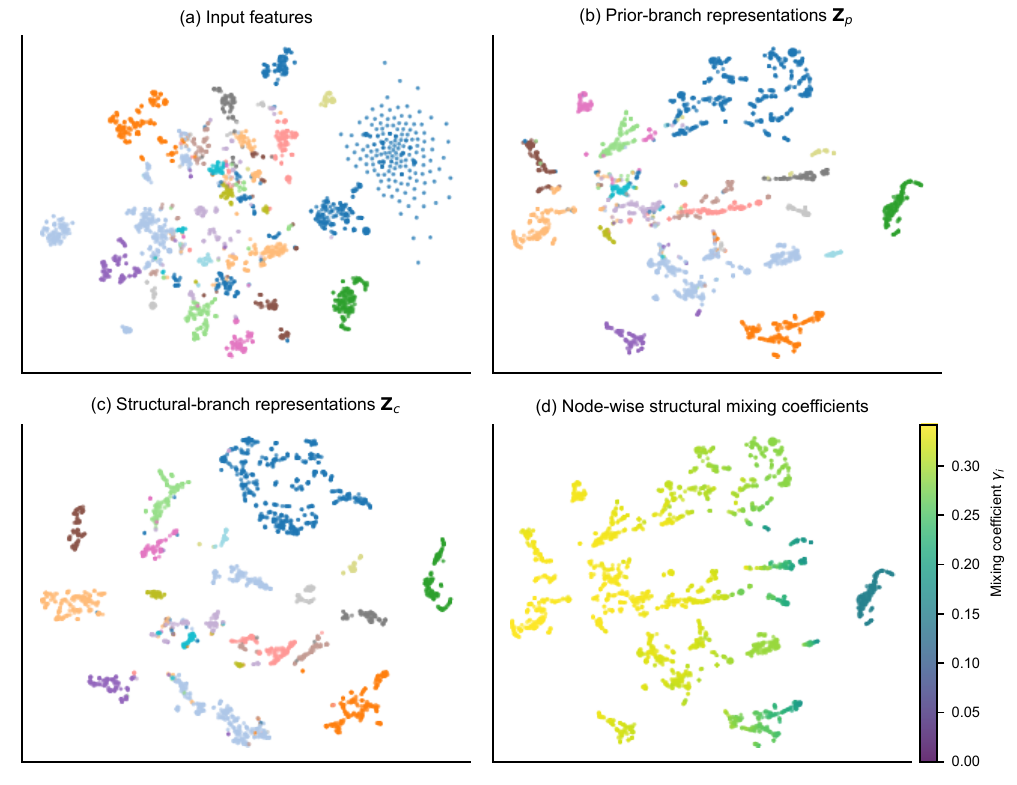} \caption{Visualization of input features, branch representations, and structural mixing coefficients on AMG18. Panels~(a)--(c) show two-dimensional projections of input features, prior-branch representations $\mathbf{Z}_p$, and structural-branch representations $\mathbf{Z}_c$, respectively, with nodes colored consistently by APT group. Panel~(d) uses the same coordinates as panel~(b), with node colors indicating the structural mixing coefficients $\gamma_i$.} \label{fig:tsne_gate} 
\end{figure*}

\subsubsection{Representation and Fusion Visualization} \label{sec:amg18_representation_visualization}

Figure~\ref{fig:tsne_gate} provides a qualitative view of the learned representations and node-adaptive fusion on AMG18. Compared with the input-feature projection in panel~(a), the projections of the prior-branch representations $\mathbf{Z}_p$ in panel~(b) and structural-branch representations $\mathbf{Z}_c$ in panel~(c) show more visually distinct clusters for several APT groups. These patterns illustrate the class-related organization of the learned branch representations.

Panel~(d) displays the node-wise structural mixing coefficients $\gamma_i$ on the same two-dimensional projection of $\mathbf{Z}_p$ used in panel~(b). The coefficients vary across nodes, including within some clusters visible in panel~(b). Together, the visualizations illustrate class-related organization in the branch representations and node-dependent weighting in prediction fusion.

\subsection{Hyperparameter Sensitivity} \label{sec:sensitivity}

We examine the sensitivity of DualGSL to the neighborhood size $k$, branch-specific classification weight $\lambda_{\mathrm{view}}$, source-node ratio, and class-anchor alignment weight $\lambda_{\mathrm{center}}$ on AMG18. Each sweep varies one hyperparameter while keeping the others at their main-experiment settings in Table~\ref{tab:hyperparameters}. Table~\ref{tab:param_sensitivity_apt} summarizes the results.

The neighborhood-size sweep achieves its highest mean accuracy (89.22\%) and Macro-F1 (87.63\%) at $k=5$. Across the evaluated values, mean Macro-F1 ranges from 86.79\% to 87.63\%. Performance does not improve monotonically as $k$ increases, indicating that enlarging the prior neighborhoods alone does not necessarily improve prediction.

All evaluated positive values of $\lambda_{\mathrm{view}}$ yield higher mean Macro-F1 than $\lambda_{\mathrm{view}}=0$. The main setting of $\lambda_{\mathrm{view}}=0.40$ improves Macro-F1 from 86.32\% to 87.28\%, a gain of 0.96 percentage points. The highest mean accuracy (89.25\%) and Macro-F1 (87.49\%) in this sweep occur at $\lambda_{\mathrm{view}}=0.10$. These results support providing direct classification supervision to both branches alongside the fused classification objective.

Performance varies comparatively little with the source-node ratio: mean Macro-F1 ranges from 86.92\% to 87.28\% as the ratio varies from 0.3 to 0.7. The default ratio of 0.5 achieves the highest mean Macro-F1, whereas 0.7 achieves the highest mean accuracy. For class-anchor alignment, $\lambda_{\mathrm{center}}=0.05$ achieves the highest mean accuracy and Macro-F1 among the evaluated weights. Its Macro-F1 of 87.28\% exceeds the result without alignment by 0.30 percentage points, supporting class-anchor alignment at the default weight.

\begin{table}[t] \centering \caption{Hyperparameter sensitivity on AMG18. Results are mean $\pm$ standard deviation over five random seeds. Bold means indicate the best result for each metric within each sweep. $^\dagger$ marks the main-experiment settings; the corresponding rows report the same five AMG18 runs as Table~\ref{tab:main_results}.} \label{tab:param_sensitivity_apt} \small \setlength{\tabcolsep}{4pt} \begin{tabular}{@{}lccc@{}} \toprule Parameter & Value & Acc. (\%) & Macro-F1 (\%) \\ \midrule $k$ & 3 & $88.65\pm0.73$ & $86.79\pm0.71$ \\ $k$ & 5 & $\mathbf{89.22}\pm0.40$ & $\mathbf{87.63}\pm0.62$ \\ $k$ & 8 & $88.53\pm0.62$ & $86.89\pm0.76$ \\ $k$ & $10^\dagger$ & $88.76\pm0.23$ & $87.28\pm0.64$ \\ $k$ & 15 & $88.95\pm0.77$ & $87.21\pm0.99$ \\ \midrule $\lambda_{\mathrm{view}}$ & 0.00 & $87.93\pm0.31$ & $86.32\pm0.70$ \\ $\lambda_{\mathrm{view}}$ & 0.10 & $\mathbf{89.25}\pm0.70$ & $\mathbf{87.49}\pm1.05$ \\ $\lambda_{\mathrm{view}}$ & 0.20 & $88.78\pm0.83$ & $87.25\pm1.00$ \\ $\lambda_{\mathrm{view}}$ & $0.40^\dagger$ & $88.76\pm0.23$ & $87.28\pm0.64$ \\ $\lambda_{\mathrm{view}}$ & 0.60 & $88.65\pm0.45$ & $86.90\pm0.82$ \\ \midrule Source-node ratio & 0.3 & $88.51\pm0.41$ & $86.92\pm0.37$ \\ Source-node ratio & $0.5^\dagger$ & $88.76\pm0.23$ & $\mathbf{87.28}\pm0.64$ \\ Source-node ratio & 0.7 & $\mathbf{88.79}\pm0.27$ & $87.09\pm0.43$ \\ \midrule $\lambda_{\mathrm{center}}$ & 0.00 & $88.59\pm0.41$ & $86.98\pm0.51$ \\ $\lambda_{\mathrm{center}}$ & 0.01 & $88.51\pm0.74$ & $86.95\pm1.27$ \\ $\lambda_{\mathrm{center}}$ & $0.05^\dagger$ & $\mathbf{88.76}\pm0.23$ & $\mathbf{87.28}\pm0.64$ \\ $\lambda_{\mathrm{center}}$ & 0.10 & $88.29\pm0.33$ & $86.54\pm0.50$ \\ \bottomrule \end{tabular} \end{table}

\section{Conclusion and Future Work} \label{sec:conclusion}

This study demonstrates the value of coordinating feature-derived priors and task-adaptive structural information at the prediction level. DualGSL preserves distinct propagation paths and learns a node-specific balance between their predictions, allowing both sources of information to contribute within a jointly trained framework. Evaluation on AMG18, 20News, and USPS yields the highest cross-dataset average accuracy and Macro-F1 among the compared methods. The analyses on AMG18 further connect structural adaptation with predictive benefits: learned connections expand the prior neighborhoods, branch ablations demonstrate the benefit of retaining both propagation paths, and fixed-checkpoint comparisons show that logit fusion improves Macro-F1 over either branch head. Together, these findings highlight prediction-level fusion as an effective mechanism for coordinating prior preservation and task-adaptive graph learning.

Future work will investigate how prior-graph quality, label availability, and branch confidence influence node-wise fusion weights and their associated predictive gains, with the aim of developing more reliable fusion strategies across data conditions. We will also extend DualGSL to inductive settings, examining how the prior and learned propagation paths can incorporate previously unseen nodes while maintaining effective prediction fusion. For larger graphs, we will explore sparse affinity learning and candidate-neighbor selection to reduce graph-construction and propagation costs while preserving the complementary information provided by the two branches.

\bibliographystyle{IEEEtran}
\bibliography{ref}

@article{song2023topoenha,
  title = {Topological enhanced graph neural networks for semi-supervised node classification},
  author = {Rui Song and Fausto Giunchiglia and Ke Zhao and Hao Xu},
  year = {2023},
  journal = {Applied Intelligence},
  doi = {10.1007/s10489-023-04739-6},
  url = {https://doi.org/10.1007/s10489-023-04739-6},
}

@article{yan2025selehop,
  title = {Selective-Hop Graph Neural Networks for semi-supervised node classification},
  author = {Hui Yan and Guoguo Ai and Xin Li},
  year = {2025},
  journal = {Neurocomputing},
  doi = {10.1016/j.neucom.2024.128812},
  url = {https://doi.org/10.1016/j.neucom.2024.128812},
}

@article{eliasof2024globloca,
  title = {Global-local graph neural networks for node-classification},
  author = {Moshe Eliasof and Eran Treister},
  year = {2024},
  journal = {Pattern Recognition Letters},
  doi = {10.1016/j.patrec.2024.06.013},
  url = {https://doi.org/10.1016/j.patrec.2024.06.013},
}

@inproceedings{wu2023nodescal,
  title = {{NodeFormer}: A Scalable Graph Structure Learning
           Transformer for Node Classification},
  author = {Qitian Wu and Wentao Zhao and Zenan Li and
            David P. Wipf and Junchi Yan},
  booktitle = {Advances in Neural Information Processing Systems},
  volume = {35},
  year = {2022},
  url = {https://proceedings.neurips.cc/paper_files/paper/2022/hash/af790b7ae573771689438bbcfc5933fe-Abstract-Conference.html}
}

@article{li2024gagsglob,
  title = {GaGSL: Global-augmented Graph Structure Learning via Graph Information Bottleneck},
  author = {Shuangjie Li and Jiangqing Song and Baoming Zhang and
          Gaoli Ruan and Junyuan Xie and Chongjun Wang},
  journal = {arXiv preprint arXiv:2411.04356},
  year = {2024},
  url = {https://arxiv.org/abs/2411.04356},
}

@article{zheng2024rethstru,
  title = {Rethinking Structure Learning For Graph Neural Networks},
  author = {Yilun Zheng and Zhuofan Zhang and Ziming Wang and
          Xiang Li and Sitao Luan and Xiaojiang Peng and Lihui Chen},
  journal = {arXiv preprint arXiv:2411.07672},
  year = {2024},
  url = {https://arxiv.org/abs/2411.07672},
}

@inproceedings{han2025unceawar,
  title = {Uncertainty-Aware Graph Structure Learning},
  author = {Shen Han and Zhiyao Zhou and Jiawei Chen and
            Zhezheng Hao and Sheng Zhou and Gang Wang and
            Yan Feng and Chun Chen and Can Wang},
  year = {2025},
  booktitle = {Proceedings of the ACM on Web Conference 2025},
  pages = {4863--4874},
  doi = {10.1145/3696410.3714927},
  url = {https://doi.org/10.1145/3696410.3714927},
}

@inproceedings{wang2025weiggrap,
  title = {Weighted Graph Structure Learning with Attention Denoising for Node Classification},
  author = {Tingting Wang and Jiaxin Su and Haobing Liu and Ruobing Jiang},
  year = {2025},
  booktitle = {2025 40th Youth Academic Annual Conference of Chinese Association of Automation (YAC)},
  pages = {42--47},
  doi = {10.1109/YAC66630.2025.11150183},
  url = {https://doi.org/10.1109/YAC66630.2025.11150183},
}

@article{han2026infograp,
  title = {Informative Graph Structure Learning},
  author = {Shen Han and Zhiyao Zhou and Jiawei Chen and Sheng Zhou and
          Canghong Jin and Hai Lin and Da Zhong Li and Bingde Hu and
          Can Wang},
  journal = {arXiv preprint arXiv:2605.16809},
  year = {2026},
  url = {https://arxiv.org/abs/2605.16809},
}

@article{luo2025labeas,
  title = {Label as Equilibrium: A performance booster for Graph Neural Networks on node classification},
  author = {Yi Luo and Guangchun Luo and Guiduo Duan and Aiguo Chen},
  year = {2025},
  journal = {Neural Networks},
  doi = {10.1016/j.neunet.2025.107284},
  url = {https://doi.org/10.1016/j.neunet.2025.107284},
}

@article{shi2025jlrgcn,
  title = {JLR-GCN: Joint label-aware and relation-aware graph convolution neural network for heterogeneous graph representations},
  author = {Zhenquan Shi and Wengjian Zhang and Jiashuang Huang and Weiping Ding},
  year = {2025},
  journal = {Information Sciences},
  doi = {10.1016/j.ins.2025.122011},
  url = {https://doi.org/10.1016/j.ins.2025.122011},
}

@article{bose2025cangrap,
  title = {Can Graph Neural Networks Tackle Heterophily? Yes, With a Label-Guided Graph Rewiring Approach!},
  author = {Kushal Bose and Saptarshi Banerjee and Swagatam Das},
  year = {2025},
  journal = {IEEE Transactions on Neural Networks and Learning Systems},
  doi = {10.1109/tnnls.2025.3565108},
  url = {https://doi.org/10.1109/tnnls.2025.3565108},
}

@inproceedings{liang2023predglob,
  title = {Predicting Global Label Relationship Matrix for Graph Neural Networks under Heterophily},
  author = {Langzhang Liang and Xiangjing Hu and Zenglin Xu and Zixing Song and Irwin King},
  year = {2023},
  booktitle = {Advances in Neural Information Processing Systems 36},
  doi = {10.52202/075280-0480},
  url = {https://doi.org/10.52202/075280-0480},
}

@inproceedings{liao2023dualdual,
  title = {DualHGNN: A Dual Hypergraph Neural Network for Semi-Supervised Node Classification based on Multi-View Learning and Density Awareness},
  author = {Jianpeng Liao and Jun Yan and Qian Tao},
  year = {2023},
  booktitle = {2023 International Joint Conference on Neural Networks (IJCNN)},
  doi = {10.1109/ijcnn54540.2023.10191471},
  url = {https://doi.org/10.1109/ijcnn54540.2023.10191471},
}

@article{han2023dualadap,
  title = {Dual adaptive learning multi-task multi-view for graph network representation learning},
  author = {Beibei Han and Yingmei Wei and Qingyong Wang and Shanshan Wan},
  year = {2023},
  journal = {Neural Networks},
  doi = {10.1016/j.neunet.2023.02.026},
  url = {https://doi.org/10.1016/j.neunet.2023.02.026},
}

@article{wang2024hetegrap,
  title = {Heterogeneous graph convolutional network for multi-view semi-supervised classification},
  author = {Shiping Wang and Sujia Huang and Zhihao Wu and Rui Liu and Yong Chen and Dell Zhang},
  year = {2024},
  journal = {Neural Networks},
  doi = {10.1016/j.neunet.2024.106438},
  url = {https://doi.org/10.1016/j.neunet.2024.106438},
}

@article{zeng2024multview,
  title = {Multi-view Heterogeneous Graph Neural Networks for Node Classification},
  author = {Xi Zeng and Fang-Yuan Lei and Chang-Dong Wang and Qing-Yun Dai},
  year = {2024},
  journal = {Data Science and Engineering},
  doi = {10.1007/s41019-024-00253-y},
  url = {https://doi.org/10.1007/s41019-024-00253-y},
}

@article{shi2025explunif,
  title = {Exploring unified cross-view hypergraph generation for multi-view semi-supervised classification},
  author = {Zhibin Shi and Zhenghong Lin and Weihong Lin and Shiping Wang},
  year = {2025},
  journal = {Neural Networks},
  doi = {10.1016/j.neunet.2025.107451},
  url = {https://doi.org/10.1016/j.neunet.2025.107451},
}

@inproceedings{kipf2017semi,
  title = {Semi-Supervised Classification with Graph Convolutional Networks},
  author = {Thomas N. Kipf and Max Welling},
  booktitle = {International Conference on Learning Representations},
  year = {2017},
  url = {https://openreview.net/forum?id=SJU4ayYgl},
}

@inproceedings{zhao2023selfsupervised,
  title = {Self-Supervised Graph Structure Refinement for Graph Neural Networks},
  author = {Jianan Zhao and Qianlong Wen and Mingxuan Ju and Chuxu Zhang and Yanfang Ye},
  booktitle = {Proceedings of the Sixteenth ACM International Conference on Web Search and Data Mining},
  year = {2023},
  pages = {159--167},
  doi = {10.1145/3539597.3570455},
  url = {https://doi.org/10.1145/3539597.3570455},
}

@inproceedings{chen2020iterative,
  title = {Iterative Deep Graph Learning for Graph Neural Networks: Better and Robust Node Embeddings},
  author = {Yu Chen and Lingfei Wu and Mohammed J. Zaki},
  booktitle = {Advances in Neural Information Processing Systems},
  volume = {33},
  pages = {19314--19326},
  year = {2020},
}

@inproceedings{huang2024elugcn,
  title     = {Enhancing the Influence of Labels on Unlabeled
               Nodes in Graph Convolutional Networks},
  author    = {Huang, Jincheng and Mo, Yujie and Shi, Xiaoshuang
               and Feng, Lei and Zhu, Xiaofeng},
  booktitle = {Proceedings of the 42nd International
               Conference on Machine Learning},
  series    = {Proceedings of Machine Learning Research},
  volume    = {267},
  pages     = {25613--25633},
  year      = {2025},
  publisher = {PMLR},
  url       = {https://proceedings.mlr.press/v267/huang25x.html}
}

@inproceedings{liu2022compact,
  title = {Compact Graph Structure Learning via Mutual Information Compression},
  author = {Nian Liu and Xiao Wang and Lingfei Wu and Yu Chen and Xiaojie Guo and Chuan Shi},
  booktitle = {Proceedings of the ACM Web Conference 2022},
  year = {2022},
  pages = {1601--1610},
  doi = {10.1145/3485447.3512206},
  url = {https://doi.org/10.1145/3485447.3512206},
}

@article{yu2023label,
  title   = {Label-Enhanced Graph Neural Network for
             Semi-Supervised Node Classification},
  author  = {Yu, Le and Sun, Leilei and Du, Bowen
             and Zhu, Tongyu and Lv, Weifeng},
  journal = {IEEE Transactions on Knowledge and Data Engineering},
  volume  = {35},
  number  = {11},
  pages   = {11529--11540},
  year    = {2023},
  doi     = {10.1109/TKDE.2022.3231660}
}

@patent{Sun2022patentFeature,
    author       = {Sun, Yuxia and Zhao, Jingjing and Weng, Jian and Lin, Song},
  title        = {Feature engineering method for author attribution of malware on the Windows platform},
  note         = "CN Patent ZL202010741380X, May. 24, 2022."
}

@patent{Sun2023patentAttr,
    author       = "Sun, Yuxia and Chen, Yanjia and Weng, Jian and Chen , Zhao",
  title        = "Open-set recognition method, apparatus, device, and medium for APT malware attribution",
  note         = "CN Patent ZL202010741373.X, Nov. 7, 2023."
}

@article{sun2024mgap,
  title={MGAP 3: Malware Group Attribution Based on PerceiverIO and Polytype Pre-Training},
  author={Sun, Yuxia and Chen, Shiqi and Lin, Song and Sun, Aoxiang and Long, Saiqin and Li, Zhetao},
  journal={IEEE Transactions on Dependable and Secure Computing},
  year={2024},
  publisher={IEEE}
}

@article{sun2026uap4ma,
  author  = {Sun, Yuxia and Hu, Hengfeng and Liu, Yepang and
             Liang, Haocheng and Chen, Shiqi and Liu, Zhiquan and
             Ma, Jianfeng},
  title   = {{UAP4MA}: Leveraging Multi-Agent Bandits to Generate
             Universal Adversarial Perturbations for Malware Attribution},
  journal = {IEEE Transactions on Dependable and Secure Computing},
  year    = {2026},
  volume  = {23},
  number  = {4},
  pages   = {8985--8997},
  doi     = {10.1109/TDSC.2026.3689685}
}

@misc{amg18dataset,
  author       = {Sun, Yuxia and et al.},
  title        = {{ARMA}: Adversarially Robust Malware Attribution
                  with Global Perturbations and Adversarial
                  Consistency Regularization},
  year         = {2026},
  howpublished = {GitHub repository},
  url          = {https://github.com/Yuxia-Sun/ARMA},
  note         = {AMG18 dataset. Accessed: Oct. 5, 2026}
}

@inproceedings{lang1995newsweeder,
  author    = {Lang, Ken},
  title     = {{NewsWeeder}: Learning to Filter Netnews},
  booktitle = {Proceedings of the 12th International Conference
               on Machine Learning},
  year      = {1995},
  pages     = {331--339},
  publisher = {Morgan Kaufmann},
  doi       = {10.1016/B978-1-55860-377-6.50048-7}
}

@article{hull1994database,
  author  = {Hull, Jonathan J.},
  title   = {A Database for Handwritten Text Recognition Research},
  journal = {IEEE Transactions on Pattern Analysis and Machine Intelligence},
  year    = {1994},
  volume  = {16},
  number  = {5},
  pages   = {550--554},
  doi     = {10.1109/34.291440}
}

@misc{aptmalware2019,
  author       = {{Cyber Research}},
  title        = {{APT} Malware Dataset},
  year         = {2019},
  howpublished = {GitHub repository},
  url          = {https://github.com/cyber-research/APTMalware}
}

@misc{virussign,
  author = {{VirusSign}},
  title  = {{VirusSign}: Open Malware Database},
  url    = {https://www.virussign.com/}
}

@misc{allmpnetbasev2,
  author       = {Reimers, Nils and Gurevych, Iryna},
  title        = {{sentence-transformers/all-mpnet-base-v2}},
  year         = {2022},
  howpublished = {Hugging Face Model Hub},
  url          = {https://huggingface.co/sentence-transformers/all-mpnet-base-v2},
}

@inproceedings{he2016deep,
  author = {He, Kaiming and Zhang, Xiangyu and
            Ren, Shaoqing and Sun, Jian},
  title = {Deep Residual Learning for Image Recognition},
  booktitle = {Proceedings of the IEEE Conference on
               Computer Vision and Pattern Recognition},
  pages = {770--778},
  year = {2016}
}
\begin{IEEEbiography}[{\includegraphics[width=1in,height=1.25in,clip,keepaspectratio]{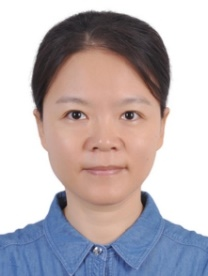}}]{Yuxia Sun}
(Member, IEEE) received her B.S. degree in Computer Science from the Huazhong University of Science and Technology, Wuhan, China, and her Ph.D. degree in Computer Science from Sun Yat-sen University, Guangzhou, China. She is currently an Associate Professor in the Department of Computer Science at Jinan University, Guangzhou, China. She previously served as a Research Associate at The University of Hong Kong and The Hong Kong Polytechnic University, Hong Kong SAR, and as a Research Scholar in the College of Computing at the Georgia Institute of Technology, Atlanta, GA, USA. Her current research interests include artificial intelligence, system security, network security, and software engineering.
\end{IEEEbiography}

\begin{IEEEbiography}[{\includegraphics[width=1in,height=1.25in,clip,keepaspectratio]{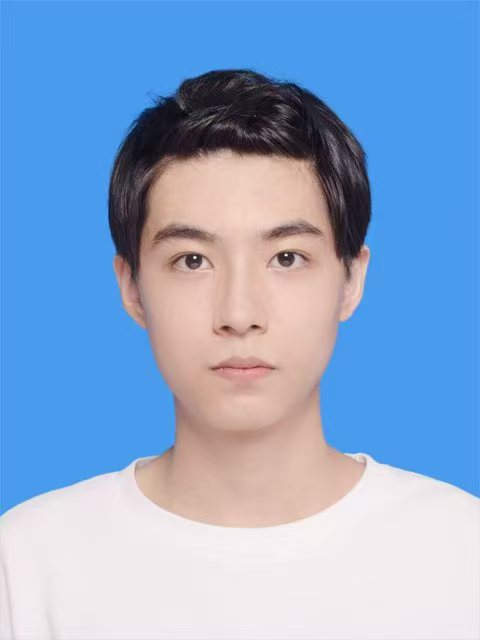}}]{Yutszyuk Wong}
is currently pursuing the Bachelor's degree with the Department of Computer Science, College of Information Science and Technology, Jinan University, Guangzhou, China. His research interests include natural language processing and large language models.
\end{IEEEbiography}

\begin{IEEEbiography}[{\includegraphics[width=1in,height=1.25in,clip,keepaspectratio]{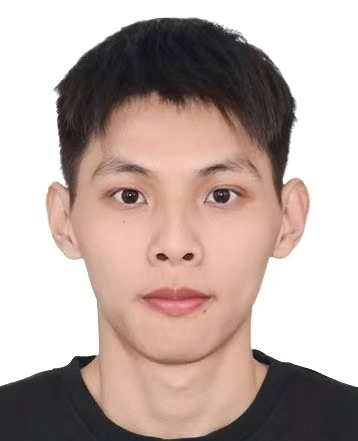}}]{Huihong Chen}
received his master's degree from Jinan University, Guangzhou, China. His research interests include machine learning and software security.
\end{IEEEbiography}

\begin{IEEEbiography}[{\includegraphics[width=1in,height=1.25in,clip,keepaspectratio]{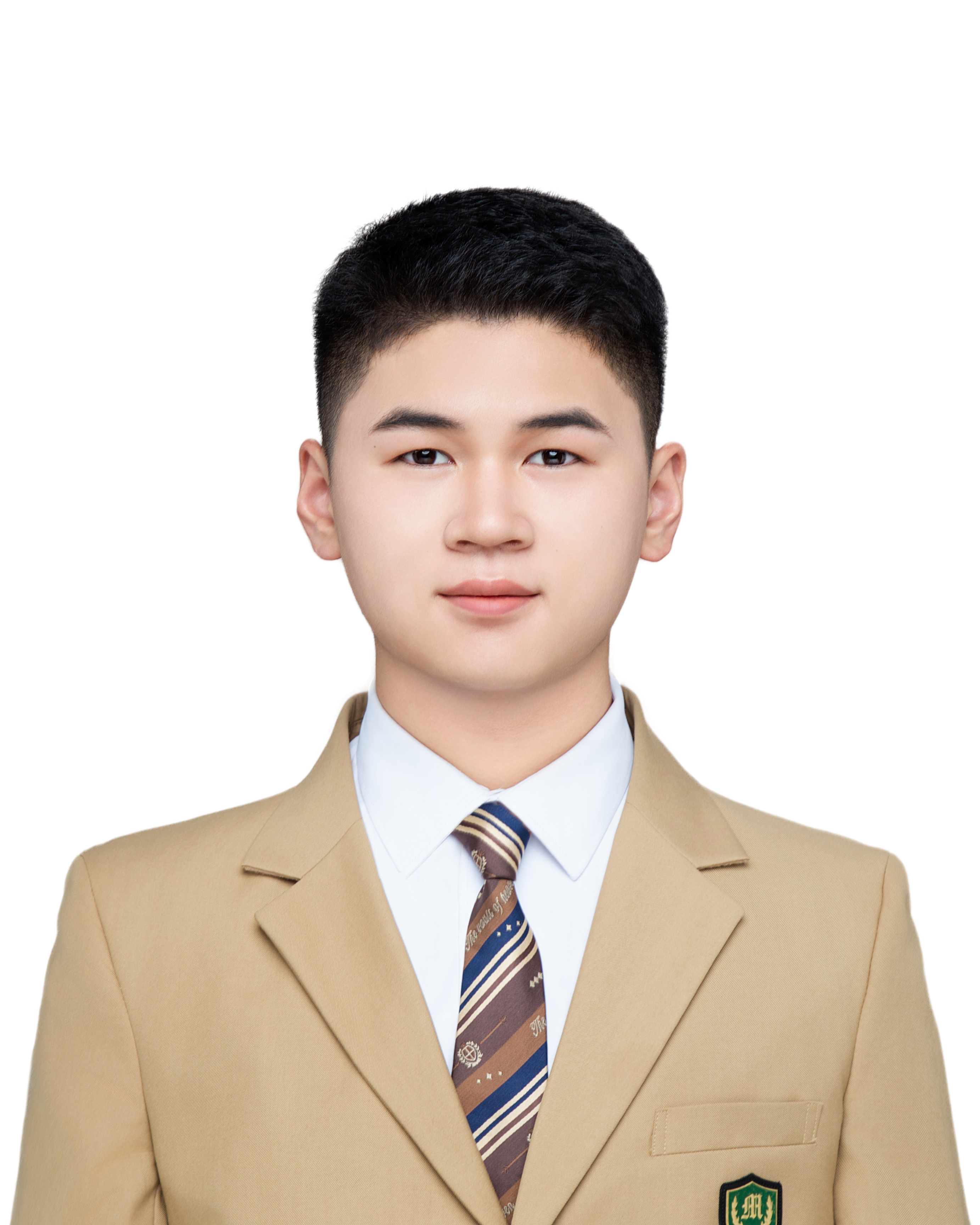}}]{Longhao Lv}
is currently pursuing the Bachelor's degree in computer science and technology with the International School, Jinan University, Guangzhou, China. His research interest is primarily concentrated on machine learning and large language models.
\end{IEEEbiography}

\begin{IEEEbiography}[{\includegraphics[width=1in,height=1.25in,clip,keepaspectratio]{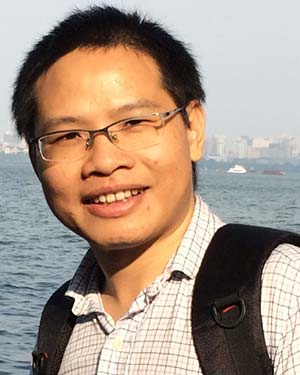}}]{Chang-Dong Wang}
(Senior Member, IEEE) received the Ph.D. degree in computer science from Sun Yat-sen University, Guangzhou, China, in 2013. He was a visiting student with the University of Illinois at Chicago from January 2012 to November 2012. He joined Sun Yat-sen University in 2013, where he is currently a Professor with the School of Computer Science and Engineering. His current research interests include machine learning and data mining. He has published more than 100 scientific papers in international journals and conferences such as IEEE Transactions on Pattern Analysis and Machine Intelligence, IEEE Transactions on Knowledge and Data Engineering, IEEE Transactions on Cybernetics, IEEE Transactions on Neural Networks and Learning Systems, ACM Transactions on Knowledge Discovery from Data, ACM Transactions on Intelligent Systems and Technology, IEEE Transactions on Systems, Man, and Cybernetics: Systems, IEEE Transactions on Industrial Informatics, KDD, AAAI, IJCAI, CVPR, ICDM, CIKM and SDM. His ICDM 2010 paper won the Honorable Mention for Best Research Paper Awards. He won 2012 Microsoft Research Fellowship Nomination Award. He was awarded 2015 Chinese Association for Artificial Intelligence (CAAI) Outstanding Dissertation. He is an associate editor in Journal of Artificial Intelligence Research (JAIR) and Neural Networks.
\end{IEEEbiography}

\begin{IEEEbiography}[{\includegraphics[width=1in,height=1.25in,clip,keepaspectratio]{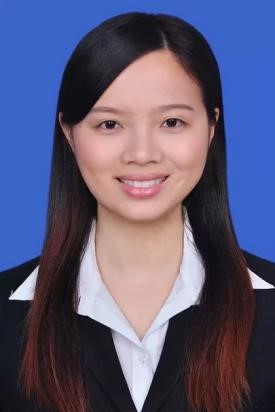}}]{Li Zeng}
received her Ph.D. degree in Mathematics from Xiangtan University in 2025. She received her B.Eng. degree in Software Engineering in 2013 and her M.Sc. degree in Computer Science and Technology in 2016, both from the same university. Her current research interests include trustworthy machine learning. She is with the School of Computer Science and Technology, Changsha University of Science and Technology, Changsha 410114, China.
\end{IEEEbiography}

\vfill

\end{document}